\pdfoutput=1
\documentclass[11pt]{article}

\usepackage{acl}
\usepackage{times}
\usepackage{latexsym}

\usepackage[T1]{fontenc}
\usepackage[utf8]{inputenc}
\usepackage{url}
\usepackage{amssymb}
\usepackage[whole]{bxcjkjatype}
\usepackage{url}
\usepackage{multirow}
\usepackage{enumitem}
\usepackage{setspace}
\usepackage{booktabs}

\newcommand{\revise}[1]{\textcolor{black}{#1}}

\title{MELD-ST: An Emotion-aware Speech Translation Dataset}

\author{
  Sirou Chen${}^{1}$\textsuperscript{\dag\ddag} \, Sakiko Yahata${}^{2}$\textsuperscript{\dag} \, Shuichiro Shimizu${}^{2}$\textsuperscript{\dag} \\
{\bf Zhengdong Yang}${}^{2}$ \, {\bf Yihang Li}${}^{3}$\textsuperscript{\ddag} \, {\bf Chenhui Chu}${}^{2}$ \, {\bf Sadao Kurohashi}${}^{2,4}$ \vspace{0.5em} \\
${}^{1}$Technical University of Munich, Germany \, ${}^{2}$Kyoto University, Japan \\
${}^{3}$SenseTime, Japan \, ${}^{4}$National Institute of Informatics, Japan \vspace{0.5em} \\
\texttt{ge23zuh@mytum.de} \, \texttt{\{yahata,sshimizu\}@nlp.ist.i.kyoto-u.ac.jp}
}

\begin{document}
\maketitle

\begingroup
\renewcommand\thefootnote{\dag}
\footnotetext{Equal contribution.}
\endgroup

\begingroup
\renewcommand\thefootnote{\ddag}
\footnotetext{Work done while at Kyoto University.}
\endgroup

\begin{abstract}
Emotion plays a crucial role in human conversation. This paper underscores the significance of considering emotion in speech translation. We present the MELD-ST dataset for the emotion-aware speech translation task, comprising English-to-Japanese and English-to-German language pairs. Each language pair includes about $10,000$ utterances annotated with emotion labels from the MELD dataset. Baseline experiments using the \textsc{SeamlessM4T} model on the dataset indicate that fine-tuning with emotion labels can enhance translation performance in some settings, highlighting the need for further research in emotion-aware speech translation systems.
\end{abstract}

\section{Introduction} \label{sec:introduction}

Human conversation naturally involves emotion.
An addressee relies on the speaker's multimodal cues, such as vocal tones and facial expressions, to understand the meaning of an utterance.
Handling emotions in machine learning systems is therefore considered an important task, as exemplified by NLP tasks such as sentiment analysis and emotion recognition in conversation \citep{fu-etal-2023-ercsurvey}.

Considering emotion in translation is also important.
For example, the phrase "Oh my God!" can express a wide range of emotions or reactions, including surprise, shock, awe, excitement, distress, etc., to convey a strong emotional response to a situation, whether positive or negative.
Because the literal translation of this wouldn't make sense in a different culture, such emotional phrases need to be translated differently depending on the emotion.
For instance, the Japanese translation of the phrase showing surprise could be "マジか! (\textit{majika})," whereas it could be "やった! (\textit{yatta})" when it shows excitement.

Emotion has been studied in machine translation (or text-to-text translation, T2TT) studies \citep{troiano-etal-2020-emotionmt}.
However, there has been little focus on emotion in speech translation (ST).
ST is a task of translating from speech to text (speech-to-text translation, S2TT) or speech (speech-to-speech translation, S2ST).
ST performance has greatly improved over the recent years with significant efforts on datasets (detailed in Section \ref{sec:related-work}) and models \citep{seamless-2023-seamlessm4t,seamless-2023-seamless,rubenstein-etal-2023-audiopalm,radford-etal-2022-whisper}.
Although a recent study by \citet{seamless-2023-seamless} focuses on emotion, further community effort is required in this domain.

To address this gap, we present the MELD-ST dataset, which consists of about $10,000$ utterances in English-to-Japanese (En-Ja) and English-to-German (En-De) language pairs, respectively.
We extract audio and subtitles from the TV series \textit{Friends}, with emotion labels for each utterance obtained from the MELD dataset \citep{poria-etal-2019-meld}.

We conduct baseline S2TT and S2ST experiments using the \textsc{SeamlessM4T} v2 model \cite{seamless-2023-seamless}.
We show that fine-tuning improves the translation performance, and using the emotion labels can enhance the performance in some settings.

\section{Related Work} \label{sec:related-work}

ST performance has significantly improved in recent years owing to the development of datasets, including S2TT datasets such as MuST-C \cite{di-gangi-etal-2019-mustc} or CoVoST 2 \cite{wang-etal-2021-covost2} datasets, as well as S2ST datasets such as CVSS \cite{jia-etal-2022-cvss} or GigaS2S \cite{chen-etal-2021-gigaspeech,ye-etal-2023-gigast,agrawal-etal-2023-iwslt2023} datasets, inter alia.
There are also ST datasets focusing on specific aspects of ST, such as gender \citep{bentivogli-etal-2020-mustshe} or dialects \cite{anastasopoulos-etal-2022-iwslt2022}, as well as specific settings such as subtitles \cite{karakanta-etal-2020-mustcinema} or cross-language dialogue \cite{shimizu-etal-2023-speechbsd}, inter alia.

A recent study by \citet{seamless-2023-seamless} investigates emotion in ST, presenting the \textsc{SeamlessExpressive} model that captures prosody and preserves vocal style.
They created mExpresso and mDRAL corpora as extensions of existing datasets \citep{nguyen-etal-2023-expresso,ward-etal-2023-dral}, as well as automatically aligned and synthetic corpora.

The MELD-ST dataset is based on the MELD dataset \citep{poria-etal-2019-meld}, an emotion recognition dataset of multimodal multi-party conversation based on the TV series \textit{Friends}.
It contains videos with English speech and is annotated with English text, sentiment\footnote{negative, neutral, and positive} and emotion\footnote{anger, disgust, fear, sadness, joy, surprise, and neutral} labels, speaker information, and timestamps based on audio for each utterance.

The key differences between the MELD-ST dataset and existing expressive ST datasets include:
1) Inclusion of emotion labels for each utterance, which can be useful for experiments and analyses.
2) Origin from a TV series in an emotionally rich environment, with translations and acted speech by professionals, making it \revise{suitable for a pilot study of} emotion-aware ST research.
3) Coverage of the En-Ja language pair, introducing unique challenges such as the need for translation content adjustments, as described in Section \ref{sec:introduction}.

\section{MELD-ST Dataset}

\begin{table}[t]
    \centering
    \small
    \begin{tabular}{lrrr}
    \toprule
         & \# utts & En speech (h) & Target speech (h) \\
    \midrule
    \multicolumn{4}{c}{En-Ja} \\
    \midrule
        Train & 8,069 & 6.4 & 6.1 \\
        Dev. & 1,008 & 0.8 & 0.5 \\
        Test & 1,008 & 0.7 & 0.8 \\
    \midrule
    \multicolumn{4}{c}{En-De} \\
    \midrule
        Train & 9,314 & 6.9 &  7.1 \\
        Dev. & 1,164 & 0.8 & 0.9 \\
        Test & 1,164 & 0.8 & 1.0 \\
    \bottomrule
    \end{tabular}
    \caption{Statistics of the MELD-ST dataset. 
    }
    \label{tab:dataset}
\end{table}

The MELD-ST dataset is constructed from translations obtained from a Blu-ray disk and emotion labels from the MELD dataset.
This section describes the construction process.
The dataset statistics are summarized in Table \ref{tab:dataset}.

\subsection{Subtitles and Timestamp Extraction}

First, we extracted Japanese and German subtitles along with the timestamps indicating when they are displayed.
We used off-the-shelf software to obtain them from a Blu-ray disk.
The timestamps were directly obtained from the files.
Because the subtitles were included as images representing the subtitle text, we used optical character recognition (OCR) tools\footnote{Ja: \url{https://github.com/hrishikeshrt/google_drive_ocr}, De: \url{https://github.com/SubtitleEdit/subtitleedit}} to extract them.

\subsection{Text Cleaning}

We cleaned the extracted subtitles by applying some heuristics.
Specifically, we excluded speaker names at the beginning of utterances, duplicated subtitles, and apparent OCR errors.

\subsection{Alignment with MELD using Timestamps} \label{sec:dataset-alignment}

The MELD dataset contains utterances along with their timestamps.
To align MELD utterances with the subtitles extracted above, we first roughly extracted the audio and further processed them for better alignment.
Specifically, we followed the following steps:
1) Find utterance candidates (i.e., utterances where there are overlaps between MELD and subtitles timestamps).
2) Extract the audio of the candidates using the timestamps of the subtitles.
3) If there are multiple utterances in the time span, apply CTC segmentation \citep{kurzinger-etal-2020-ctcsegmentation}\footnote{We used ESPnet \cite{watanabe-etal-2018-espnet} with models \texttt{espnet/kamo-naoyuki\_wsj} and \texttt{espnet/german\_commonvoice\_blstm}.} on the candidate audio to correct timestamps, and select the candidate with the longest time overlap.
\revise{More analysis on this process is provided in Appendix \ref{sec:appendix-dataset-alignment}.}

\begin{table*}[t]
  \centering
    \small
    \begin{tabular}{clccccccc}
        \toprule
        & & Anger & Disgust & Fear & Sadness & Joy & Surprise & Neutral \\
        \midrule
        \multirow{3}{*}{En-Ja} & Train & 12.18\% & 2.95\% & 2.59\%  & 7.47\%& 15.91\% & 11.35\% & 47.54\% \\
         & Dev. & 11.81\% & 2.18\% & 3.27\%  & 8.23\% & 17.46\%& 9.50\% & 47.52\% \\
         & Test & 8.43\% & 3.87\% &2.48\%&7.24\%&18.45\%&12.00\%&47.52\% \\
        \midrule
        \multirow{3}{*}{En-De} & Train &11.76\%& 2.80\% & 2.49\% &7.04\%&16.87\%&11.77\%&47.26\%\\
         & Dev. & 10.91\%&2.15\%&2.66\%&8.51\%&17.35\%&11.17\% &47.25\%\\
         & Test & 8.76\% & 3.35\% &2.75\%&7.90\%&24.14\%&11.25\%&47.25\%  \\
        \bottomrule  
    \end{tabular}
    \caption{Emotion distribution of our MELD-ST dataset.}
  \label{tab:Emotion}
\end{table*}

\subsection{Data Split}

We split the obtained utterances to train, development, and test sets.
For part of the development and test sets, further manual cleaning was applied.\footnote{We manually analyzed non-neutral utterances (i.e., utterances with emotion labels that are not ``neutral'') in detail. Therefore, most non-neutral utterances are manually checked and corrected, whereas neutral utterances are not.}
For Japanese data, the contents of the audio and the subtitles were sometimes different, due to the gap between the written and spoken style of the language.
We used Whisper \citep{radford-etal-2022-whisper} to transcribe the audio and manually corrected the errors for part of the development and test sets.
For the training set, the subtitles are used despite the style difference.

Emotion label distribution was carefully considered during the data-splitting process, with details provided in Table \ref{tab:Emotion}. Almost half of the sentences' emotions are neutral. In the rest, some of the emotion labels are more prevalent than others, like anger, joy, and surprise.

\section{Experimental Settings}

We conducted S2TT and S2ST experiments using \textsc{SeamlssM4T}. This section provides the details of the experiments and evaluation settings.

\subsection{Models for Comparison} \label{sec:models-for-comparison}
Our models were based on \textsc{SeamlessM4T} \citep{seamless-2023-seamlessm4t,seamless-2023-seamless}, which supports both S2TT and S2ST tasks.
It integrates a massively multilingual T2TT model, an unsupervised speech representation learning model, a text-to-unit encoder and decoder, and a speech resynthesis vocoder.
Its different components can be jointly optimized, effectively addressing issues related to cascaded error propagation and domain mismatch.
The \textsc{SeamlessM4T} v2 model serves as a test bed for fine-tuning and analysis, and its speech-to-speech translation covers translation from English into $35$ languages, including Japanese and German.
Because the focus of this paper is to present the MELD-ST dataset with a reasonable baseline, we conducted experiments with the medium model.\footnote{We acknowledge that the \textsc{SeamlessExpressive} or the large models might provide better scores. However, the \textsc{SeamlessExpressive} model does not support the Japanese language, and the large model requires high computational resources for fine-tuning.
}

We compare the following three settings:
\begin{itemize}
    \item \textbf{No fine-tuning}: We evaluated the \textsc{SeamlessM4T} v2 medium model on the test set of the MELD-ST dataset.
    \item \textbf{Fine-tuning without emotion labels}: We fine-tuned the \textsc{SeamlessM4T} v2 medium model on the MELD-ST dataset without utilizing the emotion labels. 
    \item \textbf{Fine-tuning with emotion labels}: We used the emotion labels annotated for each utterance in the original MELD dataset. Following the method of \citet{gaido-etal-2020-breeding}, a study that investigated the usage of gender information for speech translation, we prepended the gold emotion labels at the beginning of the decoder input sequence during training.\footnote{Instead of introducing special tokens, the labels were prepended as text.} During testing, the emotion labels were predicted along with the translations.
\end{itemize}

The fine-tuning settings included a batch size of $4$, evaluation steps of $1,000$, and a maximum of $200$ epochs. Fine-tuning was conducted on a single Nvidia A100 80GB GPU. The training process stopped when the loss didn't improve for $10$ epochs, and the best checkpoint \revise{based on the translation quality of the development set} would be used. Each fine-tuning process lasted about $2$ hours on one GPU.

The model was fine-tuned using three data settings: En-Ja, En-De, and a mixed dataset combining En-Ja and En-De of the MELD-ST dataset.
We fine-tuned the model multiple times with the same data, resulting in different checkpoints. We report the best results obtained from these checkpoints.

\subsection{S2TT Evaluation}
The target text that is generated along with the target speech from the S2ST translation mode of SeamlessM4T, was used to compare it with the target language reference for evaluation. Instead of conventional evaluation methods like BLEU \cite{papineni-etal-2002-bleu}, BLEURT \cite{sellam-etal-2020-bleurt} was used to access translation quality because professional translations always differ greatly from literal interpretations, especially in languages like Japanese. Relying solely on n-gram matching for evaluation becomes challenging in such cases.

\subsection{S2ST Evaluation}

ASR-BLEU \citep{lee-etal-2022-directs2st} was used to evaluate the quality of the generated target speech.
We used the implementation in the Seamless Communication repository,\footnote{\url{https://github.com/facebookresearch/seamless_communication/tree/main/src/seamless_communication/cli/eval_utils}} which uses Whisper \cite{radford-etal-2022-whisper} as the underlying model.

Additionally, the generated speeches were evaluated against the original source language speech files considering various criteria such as prosody, voice similarity, pauses, and speech rate, following \citet{seamless-2023-seamless}. Details of the prosody evaluation are provided in Appendix \ref{sec:appendix-prosody}.

\section{Results} \label{sec:results}

\begin{table}[t]
  \centering
    \small
    \begin{tabular}{clrr}
        \toprule
        \multirow{2}{*}{Training Data} & \multirow{2}{*}{Fine-tuning Setting} & \multicolumn{2}{c}{Evaluation Data}  \\ 
         & & En-Ja & En-De \\
        \midrule
        - & No fine-tuning & 30.28 & 50.47 \\
        \midrule
        \multirow{2}{*}{En-Ja} & w/o emotion labels & 30.77 & - \\
         & w/ emotion labels & \bf{33.18}${}^{\dag}$ & - \\
        \midrule
        \multirow{2}{*}{En-De} & w/o emotion labels & - & 54.92 \\
         & w/ emotion labels & - & 55.13 \\
        \midrule
        \multirow{2}{*}{Mixed} & w/o emotion labels & 32.51 & 55.60 \\
         & w/ emotion labels & 32.52 & \bf{55.84} \\ 
        \bottomrule  
    \end{tabular}
    \caption{BLEURT scores in percentage on the MELD-ST test sets for the S2TT experiments with different training data and fine-tuning settings. \revise{$\dag$ indicates the difference in the scores of with and without emotion labels is statistically significant at $p < 0.05$.}}
  \label{tab:s2tt-results}
\end{table}

\begin{table}[t]
  \centering
    \small
    \begin{tabular}{clrr}
        \toprule
        \multirow{2}{*}{Training Data} & \multirow{2}{*}{Fine-tuning Setting} & \multicolumn{2}{c}{Evaluation Data}  \\ 
         & & En-Ja & En-De \\
        \midrule
        - & No fine-tuning & 0.15&8.32 \\
        \midrule
        \multirow{2}{*}{En-Ja} & w/o emotion labels  & 0.46 & - \\
         & w/ emotion labels  & 0.16 & - \\
        \midrule
        \multirow{2}{*}{En-De} & w/o emotion labels   &- & 8.37 \\
         & w/ emotion labels  &- & 8.23 \\
        \midrule
        \multirow{2}{*}{Mixed} & w/o emotion labels  & \bf{0.47}& \bf{9.82} \\
         & w/ emotion labels  & 0.14& 8.85 \\ 
        \bottomrule 
    \end{tabular}
    \caption{ASR-BLEU scores on the MELD-ST test sets for the S2ST experiments with different training data and fine-tuning settings.}
  \label{tab:s2st-results}
\end{table}

\subsection{S2TT Resuts}
Table \ref{tab:s2tt-results} shows the S2TT results.
We can see that the quality of the translations generally improved after fine-tuning, and incorporating emotion labels led to slight enhancements.
Using separate data or mixed data for fine-tuning does not show a significant difference.

\subsection{S2ST Results}

Table \ref{tab:s2st-results} shows the S2ST results. We can see that fine-tuning the \textsc{SeamlessM4T} model improves the ASR-BLEU results. However, fine-tuning with emotion labels does not help.
The rest of the metrics, such as prosody similarity and vocal similarity, do not change significantly after fine-tuning.\footnote{The scores are presented in Appendix \ref{sec:appendix-prosody} Tables \ref{tab:s2st-enja-prosody-results} and \ref{tab:s2st-ende-prosody-results}.} The reason for this is that \textsc{SeamlessM4T} doesn't bother to learn the pronunciation features in the original speech.\footnote{\revise{This means that \textsc{SeamlessM4T} does not capture the pronunciation features like prosody or vocal style in the original speech as \textsc{SeamlessExpressive} does.}} Pauses and speed changed a bit after fine-tuning, which can be assumed to be because the translation is closer to the reference after fine-tuning.

\subsection{Discussion} \label{sec:discussion}

It is generally observed that En-De provides higher translation scores compared to En-Ja in both S2TT and S2ST.
Japanese and English are very different, making the translation difficult. With the help of emotion labels, the BLEURT scores improved slightly, but not enough to be regarded as a translation with good quality.
German is more similar to English and gets higher scores. After manually checking the results, most of the sentences were very clear and correct. The reason why the addition of emotion labels does not improve the results is probably because the sentences in the test set do not change much due to the difference in emotion labels.

\section{Conclusion}
In this study, we presented the MELD-ST dataset, an ST dataset in an emotionally rich situation, which contains both En-Ja and En-De language pairs.
We conducted baseline S2TT and S2ST experiments with and without utilizing emotion labels, which showed that emotion labels can boost performance in some settings.

For improving the translation performance, several approaches can be considered, such as training a multitask model of speech emotion recognition and ST, and utilizing dialogue context in translation.

\section*{Limitations}
Some audio files in the MELD-ST dataset may contain more words than its presented text due to alignment issues. When evaluating translations of the same text with different emotion labels, it's challenging to determine the cause of differences in results. Pinpointing whether the variance stems from the emotion label or the extra information within the audio proves difficult.

The MELD-ST dataset is constructed based on acted speech, and further research is required for more natural settings such as spontaneous dialogues.
As explained in Section \ref{sec:models-for-comparison}, the models used for the experiments in this paper are basic ST models, and further performance gain could be obtained from models specifically tailored for emotion-aware speech translation.

\section*{Ethics Statements}

The MELD-ST dataset will be released to the research community with restricted access to facilitate the advancement of emotion-aware speech translation, considering the risk of unintended usage of the dataset violating the copyright.
In the dataset, English text, speech, and emotion labels were gathered from publicly available sources.
The English data used to correct timestamps, as well as the Japanese and German text and speech, were sourced from a Blu-ray disk.
Individuals seeking access to this dataset will be requested to confirm that their purpose for using it is solely for research.

Some utterances in the dataset may contain offensive contents like swear words, to the extend that is accepted by the public (i.e., to be able to appear in a TV series).

\section*{Acknowledgements}
\revise{This work was supported by JSPS KAKENHI Grant Number JP23K28144 and JP23KJ1356.}

\newpage

\bibliography{acl2024}
\bibliographystyle{acl_natbib}

\newpage
\appendix

\section{Dataset Alignment Quality} \label{sec:appendix-dataset-alignment}

\revise{
We manually checked the alignment quality with the alignment process in Section \ref{sec:dataset-alignment} for the En-Ja part of the dataset. We sampled $307$ utterances from the utterances and checked the alignment with the following criteria:
}
\begin{itemize}
    \item Correct: The English and Japanese utterances match
    \item No translation: The English utterance is not aligned with the Japanese utterance for various reasons
    \item Rough segmentation: Multiple English utterances correspond to one Japanese utterance
    \item Others: The English utterance and the Japanese utterances do not match for other reasons
\end{itemize}
\revise{
We found that $64.5\%$ belong to ``Correct,'' $16.6\%$ belong to ``No translation,'' $5.9\%$ belong to ``Rough segmentation,'' and $13.1\%$ belong to ``Others.'' For ``No translation'' the reasons are as follows: 1) The Japanese subtitles often omit information; 2) The translation does not appear as subtitles for simple utterances such as ``Hi''; 3) In cases where one Japanese utterance corresponds to multiple English utterances, only one English utterance could be aligned to the Japanese utterance. Because these parts cannot be used in the translation experiment, such utterances were automatically detected and discarded.
}

\section{S2ST Evaluation on Prosody} \label{sec:appendix-prosody}

Here, we provide details of the S2ST evaluation on prosody.
We used the \textsc{Stopes} library for the evaluation.\footnote{\url{https://github.com/facebookresearch/stopes/tree/main/stopes/eval}}
Prosody similarity, measured by \textsc{AutoPCP}, evaluates speech patterns, including rhythm, intonation, and stress. Vocal similarity, analyzed through cosine similarity with the function \textsc{VSim}, quantifies acoustic characteristics like pitch and tone. Pauses in speech and speech rate are evaluated using local\_prosody. This tool aligns audio with its corresponding text, annotates word duration, and identifies pause locations to calculate and evaluate the data.

The evaluation results on the prosody of the generated speech in the S2ST experiments are presented in Tables \ref{tab:s2st-enja-prosody-results} and \ref{tab:s2st-ende-prosody-results}.

\begin{table*}[t]
  \centering
    \begin{tabular}{clrrrr}
        \toprule
        Training Data & Fine-tuning Setting & \textsc{AutoPCP} & VSim & Pause & Rate\\
        \midrule
        - & No fine-tuning & 1.75& 0.0034 & 0.501 &-0.09 \\
        \midrule
        \multirow{2}{*}{En-Ja} & w/o emotion labels & 1.83& -0.0004 & -0.086& 0.47 \\
         & w/ emotion labels & 1.94 & -0.0020 & -0.122& 0.50 \\
        \midrule
        \multirow{2}{*}{Mixed} & w/o emotion labels & 1.88& 0.0020 &0.620& -0.12 \\
         & w/ emotion labels & 1.89 & -0.0023 &0.482& 0.08 \\ 
        \bottomrule  
    \end{tabular}
    \caption{Generated Japanese target speech evaluation results.}
  \label{tab:s2st-enja-prosody-results}
\end{table*}

\begin{table*}[t]
  \centering
    \begin{tabular}{clrrrr}
        \toprule
        Training Data & Fine-tuning Setting & \textsc{AutoPCP} & VSim & Pause & Rate\\
        \midrule
        - & No fine-tuning & 2.00& 0.0091 & 0.501 &0.09 \\
        \midrule
        \multirow{2}{*}{En-De} & w/o emotion labels & 2.07& -0.0083 & 0.091& 0.63 \\
         & w/ emotion labels & 2.05 & 0.0089 & 0.138& 0.63 \\
        \midrule
        \multirow{2}{*}{Mixed} & w/o emotion labels & 2.08& 0.0082 &0.477& 0.07 \\
         & w/ emotion labels & 2.07 & 0.0085 &0.482& 0.08 \\ 
        \bottomrule  
    \end{tabular}
    \caption{Generated German target speech evaluation results.}
  \label{tab:s2st-ende-prosody-results}
\end{table*}

\section{Dataset Examples} \label{sec:appendix-dataset-examples}
Tables \ref{tab:dataset-example-enja} and \ref{tab:dataset-example-ende} show some examples from the MELD-ST dataset.
\begin{table*}[ht]
  \centering
    \begin{tabular}{p{1.5cm}p{7cm}p{6cm}}
      \toprule
      Emotion & English & Japanese \\ \midrule
      neutral  & But um, I don't think it's anything serious. & 大したことない \\
      surprise  & Oh my God! & ヤダマジ？ウソ \\ 
      surprise  & Oh my God! & やったわ! \\
      surprise  & This sounds like a hernia. You have to-you—you go to the doctor! & ヘルニアだな医者へ \\
      joy  & Thank you…we're so excited & ありがとう楽しみです \\ 
      anger  & Hey, Ross!!! I told you I don't! & ロスいい加減にして \\ 
      \bottomrule
    \end{tabular}
    \caption{Example utterances from the MELD-ST En-Ja set.}
    \label{tab:dataset-example-enja}
\end{table*}

\begin{table*}[ht]
  \centering
    \begin{tabular}{p{1.5cm}p{7cm}p{6cm}}
      \toprule
      Emotion & English & German \\ \midrule
      neutral  & What do you mean? & Wie meinst du das? \\
      surprise  & Are you serious?&Das kann nicht sein. \\ 
      surprise  & Oh my God! & Ah! Oh, mein Gott! \\ 
      surprise  & Oh my God! & Ich glaub's nicht! \\
      joy  & Oh my God! & Ich glaub's nicht! \\ 
      joy  & Oh crap!& So ein Käse! \\ 
      joy  & They taste so good.&Die sind wirklich köstlich. \\       
      anger  & He does not look happy.&Er scheint nicht begeistert zu sein. \\
      anger  & I can't believe this! This is like the worst night ever!&Das ist wirklich der schrecklichste Abend, den ich je hatte. \\ 
      \bottomrule
    \end{tabular}
    \caption{Example utterances from the MELD-ST En-De set.}
    \label{tab:dataset-example-ende}
\end{table*}

\section{Translation Examples} \label{sec:appendix-generated-translation-examples}

Table \ref{tab:sentence-example} shows some observed examples from the S2TT experiments, which can show the potential of emotion labels to help improve translation quality.
For the first sentence, when the model is trained without considering emotion, it translates the source language directly. However, when the emotional label is incorporated, the translated text exhibits more joy.
The original text succinctly conveys the emotions without ambiguity or implication, leading to translations that remain consistent whether fine-tuned with or without the use of emotion labels.

\begin{table*}[ht]
    \begin{tabular}{p{2.5cm}p{3.0cm}p{3.0cm}p{3.0cm}p{3.0cm}}
      \toprule
      Input & Reference & No fine-tuning & w/o emotion labels &  w/ emotion labels \\ \midrule
      This game is kind of fun. & Hey, das Spiel macht doch Spaß. 
      
      (Hey, the game does make fun) & Hey, das Spiel macht doch Spaß. 
      
      (The game is a bit of fun).&Das Spiel ist ein bisschen lustig. 
      
      (The game is a bit of fun.) &Das Spiel ist ja wirklich lustig. 
      
      (The game is really fun.)\\

      \midrule
      I’m very glad you’re here. & Dass du da bist, macht mich sehr glücklich. 
      
      (It makes me very happy that you're here) &Ich bin sehr froh, dass du hier bist.

      (I'm very happy, that you are here)      
      
      & Ich bin froh, dass du da bist.
      
      (I'm happy, that you are here).&Ich bin froh, dass du da bist.
      
       (I'm happy, that you are here). \\
       \midrule
      You are so sweet. & うれしいわ
      
      (So happy) &あなたはとても可愛い。

      (You are really cute.)      
      & 優しいわ
      
      (Gentle)&かわいい人ね
      
       (Cute person) \\
      \bottomrule
    \end{tabular}
    \caption{Example of emotional fine-tuning in translation.}
    \label{tab:sentence-example}
\end{table*}

\end{document}